\def\year{2021}\relax
\documentclass[letterpaper]{article} % DO NOT CHANGE THIS
\usepackage{aaai21}  % DO NOT CHANGE THIS
\usepackage{times}  % DO NOT CHANGE THIS
\usepackage{helvet} % DO NOT CHANGE THIS
\usepackage{courier}  % DO NOT CHANGE THIS
\usepackage[hyphens]{url}  % DO NOT CHANGE THIS
\usepackage{graphicx} % DO NOT CHANGE THIS
\usepackage{natbib}  % DO NOT CHANGE THIS AND DO NOT ADD ANY OPTIONS TO IT
\usepackage{caption} % DO NOT CHANGE THIS AND DO NOT ADD ANY OPTIONS TO IT
\usepackage{color}

\usepackage{booktabs}
\usepackage{amsmath}
\usepackage{amsfonts}
\usepackage{multirow}
\usepackage{array}
\usepackage{tablefootnote}

\newcommand{\newchange}[1]{{\color{black}#1}}

\title{ACDNet: Adaptively Combined Dilated Convolution for Monocular Panorama Depth Estimation}

\author {
    Chuanqing Zhuang\textsuperscript{\rm 1}\footnote{Joint first authors with equal contribution},
    Zhengda Lu\textsuperscript{\rm 1}\footnotemark[1],
    Yiqun Wang\textsuperscript{\rm 2,3},
    Jun Xiao\textsuperscript{\rm 1}\footnote{Corresponding author},
    Ying Wang\textsuperscript{\rm 1}\\
}

\affiliations {
    \textsuperscript{\rm 1} School of Artificial Intelligence, University of Chinese Academy of Sciences\\
    \textsuperscript{\rm 2} College of Computer Science, Chongqing University \\
    \textsuperscript{\rm 3} KAUST\\
    zhuangchuanqing19@mails.ucas.ac.cn,
    \{luzhengda,xiaojun,ywang\}@ucas.ac.cn,
    yiqun.wang@cqu.edu.cn
    
}
\begin{document}

\maketitle

\begin{abstract}
Depth estimation is a crucial step for 3D reconstruction with panorama images in recent years. Panorama images maintain the complete spatial information but introduce distortion with equirectangular projection. In this paper, we propose an ACDNet based on the adaptively combined dilated convolution to predict the dense depth map for a monocular panoramic image. Specifically, we combine the convolution kernels with different dilations to extend the receptive field in the equirectangular projection. Meanwhile, we introduce an adaptive channel-wise fusion module to summarize the feature maps and get diverse attention areas in the receptive field along the channels. Due to the utilization of channel-wise attention in constructing the adaptive channel-wise fusion module, the network can capture and leverage the cross-channel contextual information efficiently. Finally, we conduct depth estimation experiments on three datasets (both virtual and real-world) and the experimental results demonstrate that our proposed ACDNet substantially outperforms the current state-of-the-art (SOTA) methods. Our codes and model parameters are accessed in \textit{https://github.com/zcq15/ACDNet}.

\end{abstract}

\section{Introduction}

The panoramic camera is a new type of camera to capture images with $180^\circ\times360^\circ$ field of view (FoV), which is convenient to obtain omnidirectional spatial information in a single shot without the post-calibration and stitching. 
With its wide usage in the fields such as virtual reality (VR) and security monitoring in recent years,
panorama depth estimation is a crucial step in a variety of downstream applications, such as semantic segmentation, layout recovery, and 3D reconstruction, to name a few. 

Generally, panorama images are represented as images on the sphere grid for warp and weft by the equirectangular projection (ERP). However, the geometric structure in the higher latitude areas is distorted since the spatial sampling rate changes with latitude.
Therefore, accurate depth estimation is difficult with conventional convolution networks in these areas.

Early works \cite{DBLP:journals/corr/abs-1709-04893,DBLP:conf/iclr/CohenGKW18} define the spherical CNNs to process the spherical signals but cause the high resource expenditure. 
And some others \cite{DBLP:conf/nips/SuG17,DBLP:conf/eccv/TatenoNT18,DBLP:conf/eccv/CoorsCG18,DBLP:journals/ral/Fernandez-Labrador20} propose the different custom convolutions to deform the convolution kernels according to the geometric structures in the 3D space of the ERP coordinates. 
These adaptive convolution kernels expand the receptive fields near the poles according to the corresponding latitude coordinates.
However, these methods still have a great potential to be further improved.

\begin{figure}[tb]
\centering
  \includegraphics[width=0.45\textwidth]{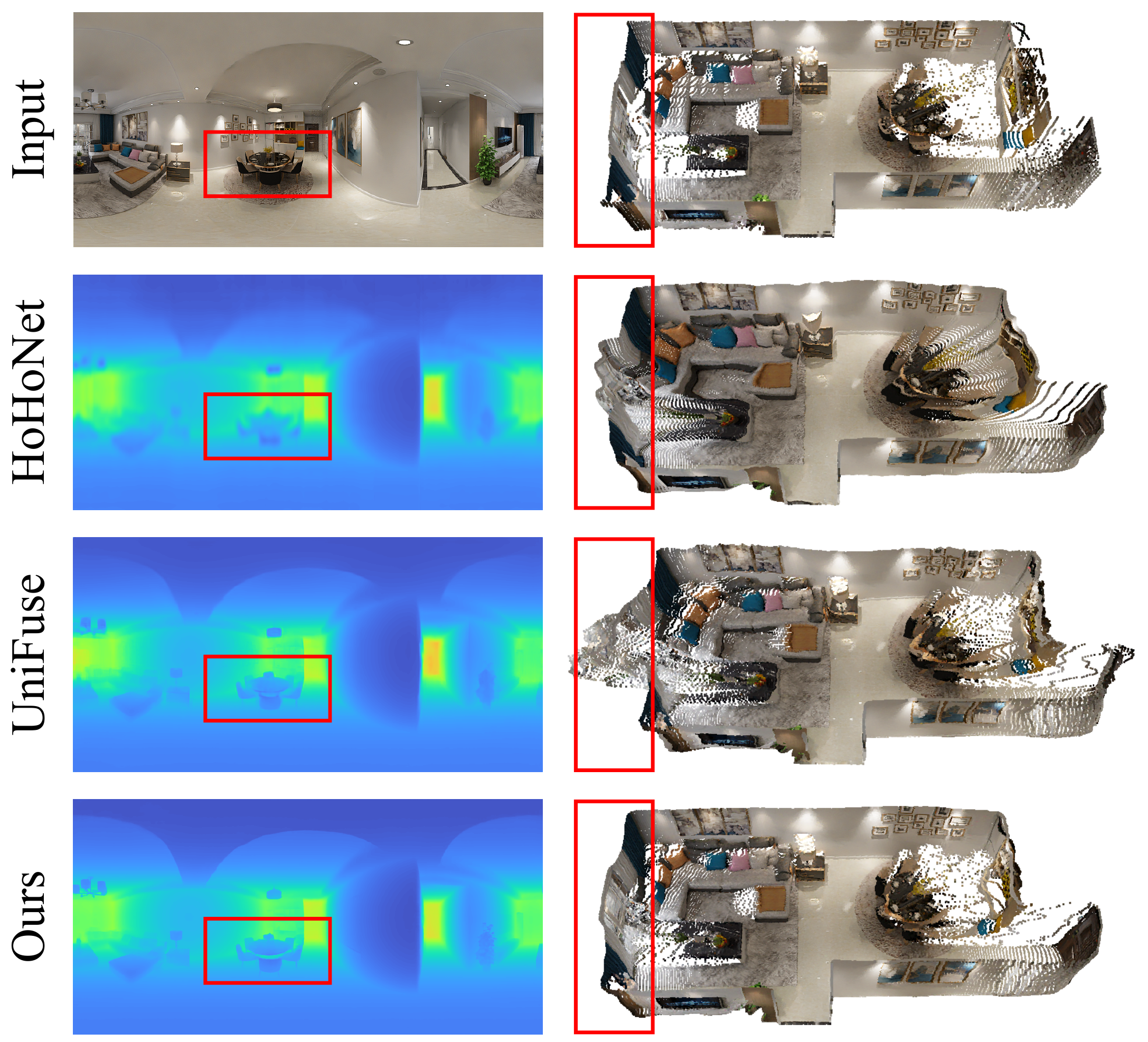}
  \caption{Reconstructed results of different models. Left: the panorama color image and the predicted depth maps. Right: point clouds generated from the ground-truth and predicted depth maps with inverse equirectangular projection. Note that our predicted depth map has more clear edges and also generates a more accurate point cloud in both overall shape and object details.}
  \label{fig:teaser}
\end{figure}

On the other hand, BiFuse \cite{DBLP:conf/cvpr/WangYSCT20} and UniFuse \cite{DBLP:journals/ral/JiangSZDH21} project the ERP image to the cubemap images to solve the distortion with the perspective projection. However, due to the limitation of the FoV in the cubemap branch, the overall layout in the reconstructed scenes can not be well restored.
Besides, HoHoNet \cite{Sun_2021_CVPR} and SliceNet \cite{Pintore_2021_CVPR} extract the horizontal 1D feature maps from gravity-aligned equirectangular projection and recover the dense 2D predictions. However, it is hard to recover the details in the columns from 1D features (see Fig.\ref{fig:teaser}). Thus, the balance between quantitative results and visual effects still needs to be considered.

In this paper, we propose the ACDNet based on the adaptively combined dilated convolution for the panoramic monocular depth estimation. 
We combine the convolution kernels with different dilations to extend the receptive field in the equirectangular projection. 
Meanwhile, we use an adaptive channel-wise fusion module to summarize the feature maps and get diverse attention areas in the receptive field along different channels.
Different from methods \cite{DBLP:conf/nips/SuG17,DBLP:conf/eccv/TatenoNT18,DBLP:conf/eccv/CoorsCG18,DBLP:journals/ral/Fernandez-Labrador20} that calculate the shapes of convolution kernels according to the latitude coordinates, we learn the focused areas in different feature channels that help the network to capture the cross-channel contextual information. 
Finally, we evaluate our method on both virtual and real-world panoramic RGB-D datasets. The experimental results show that our ACDNet and the adaptively combined dilated convolution outperform the current state-of-the-art methods.

In summary, the main contributions of this work can be summarized as follows:
\begin{itemize}
    \item[1.] We propose the adaptively combined dilated convolution to process the panorama images for monocular depth estimation, and it can be easily embedded into convolution networks by replacing the regular convolution.
    \item[2.] The interest areas can be obtained by learning different attention scores in different channels, which is more suitable for panoramic images than explicitly deforming convolution kernels in different latitudes.
    \item[3.] We perform the monocular panorama depth estimation experiments on both virtual and real-world RGB-D panorama datasets, which outperforms the SOTA methods in both quantitative metrics and visual effects.
\end{itemize}

\section{Related Work}

In this section, we describe the overview of researches on panorama depth estimation and simply introduce the applications of dilated convolution in CNNs. 

\begin{figure*}[tb]
\centering
  \includegraphics[width=0.9\textwidth]{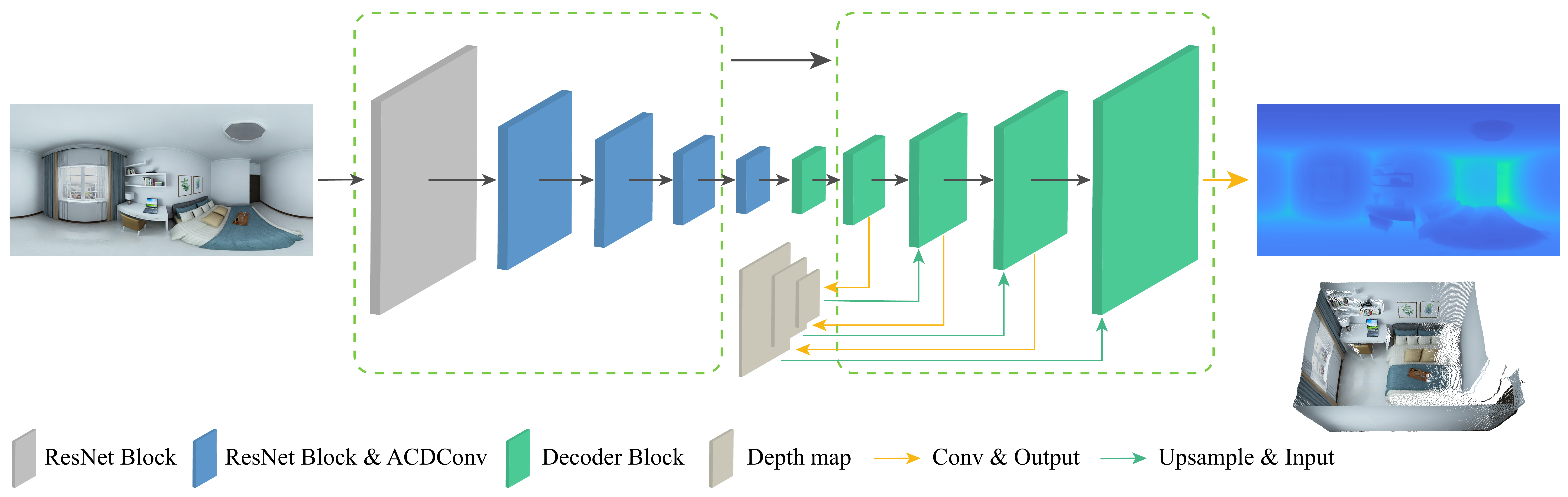}
  \caption{The architecture of our ACDNet.}
  \label{fig:network}
\end{figure*}

\subsection{Panorama Depth Estimation}

Depth estimation is an important step for 3D reconstruction, and panorama images can capture the omnidirectional spatial information for the global structure, which conduces to recover the depth in areas with weak textures.

OmniDepth \cite{DBLP:conf/eccv/ZioulisKZD18} first proposes the RectNet to estimate the depth map with a single panorama image and shows better performance than individually processing the different views of cubemap projection (CMP). 
However, this method is limited by the distortion of the geometric structures and the decrease of the FoV near the poles for panoramic images in the equirectangular projection. There are some main types of existing methods to solve this problem.

Firstly, some methods ~\cite{DBLP:conf/eccv/CoorsCG18,DBLP:conf/eccv/TatenoNT18,DBLP:journals/ral/Fernandez-Labrador20,DBLP:journals/corr/abs-1906-11096} deform the convolution kernels to adaptively extend the receptive fields of custom convolutions. Specifically, SphereNet \cite{DBLP:conf/eccv/CoorsCG18} and DistConv \cite{DBLP:conf/eccv/TatenoNT18} calculate the sampling positions for the convolution kernels with inverse gnomonic projection, and CFL \cite{DBLP:journals/ral/Fernandez-Labrador20} defines the convolution over the field of view on the spherical surface with longitudinal and latitudinal angles.
Besides, mapped convolution ~\cite{DBLP:journals/corr/abs-1906-11096} proposes a more general method to process images of any structured representation by accepting the corresponding mapping function.
These methods all produce the different convolution kernels with the change of latitude coordinates in the ERP.

Secondly, some other methods introduce the additional CMP branch with perspective projection into the network. BiFuse uses their proposed bi-projection fusion module to fuse the feature maps in two complete encoder-decoder branches. Furthermore, UniFuse removes the CMP decoder branch and proposes a more effective unidirectional fusion module. However, due to the limitation of the FoV, the CMP branch can not extract good features from areas with weak textures, e.g., the ceilings and the floors, which cripples the ability for fused features to express the spatial structures.
Last, recent works HoHoNet and SliceNet compress the 2D features to 1D features from gravity-aligned panoramic images, then they apply the RNNs to capture the global context information.

There are also still other methods to estimate panoramic depth with different strategies, including deformable convolution kernels \cite{DBLP:conf/icra/ChengWZGY20,DBLP:journals/spl/ChenLFLCG21}, geometric guidance \cite{DBLP:conf/3dim/EderMG19,DBLP:conf/cvpr/JinXZZTXYG20,DBLP:conf/eccv/ZengKG20}, self-supervised/unsupervised learning \cite{DBLP:conf/3dim/ZioulisKZAD19,DBLP:conf/itsc/ZhouWY20}, and stereo matching \cite{DBLP:conf/icra/WangSTCS20}.
These existing methods have achieved good results on depth estimation with panoramic cameras. But there is still much room for improvement in terms of quantitative results or visual effects.

\subsection{Dilated Convolution}
It is proved that dilated convolution is an effective tool for the increment of the receptive field without additional parameters and down-sampling. Apart from semantic segmentation \cite{DBLP:journals/pami/ChenPKMY18} and object detection \cite{DBLP:conf/eccv/LiuAESRFB16}, dilated convolution is also widely utilized on some other tasks such as depth estimation.

Earlier works \cite{DBLP:conf/cvpr/YuKF17,DBLP:conf/pcm/MaDWZL18} simply embed the dilated convolutions in the network. And many works use the atrous spatial pyramid pooling (ASPP) \cite{DBLP:journals/pami/ChenPKMY18} or a similar module to model contextual information. Some works \cite{DBLP:conf/cvpr/FuGWBT18,DBLP:conf/wacv/FangCCG20,DBLP:conf/eccv/ZhangYWCW20,DBLP:journals/corr/abs-2106-08615} introduce ASPP to aggregate multi-scale contextual information for better monocular depth estimation. Besides, MSDC-Net \cite{DBLP:journals/access/TianZRWHH19} combines the res-block module with different dilated rates to build the irregular shape ResNet \cite{DBLP:conf/cvpr/HeZRS16} module. CrossGuidance \cite{DBLP:journals/access/LeeLKK20} proposes a residual atrous spatial pyramid (RASP) block to analyze the large input images. And MAPUnet \cite{DBLP:journals/access/YangWZZSY21} develops the multi-layers DenseASPP with more scales to cover more pixels. 

Different from modules similar to ASPP, we combine the dilated convolutions as an equivalent large kernel convolution and apply it to replace the regular convolution layers in ResNet blocks. This operator enlarges the receptive field and produces a variety of interest areas in the receptive field.

\section{Approach}

For the reconstruction of the indoor scene with a single panoramic image, we propose the ACDNet with adaptively combined dilated convolution (ACDConv) layers to estimate the depth map.
In the following text, we first present the architecture of the ACDNet, then we show the implementation of the ACDConv to extract feature maps from panorama images. Finally, we introduce the loss function in our approach.

\subsection{Architecture}

We propose the ACDNet to estimate the depth map with a single panorama image as illustrated in Fig.\ref{fig:network}. In general, the ACDNet is a conventional network with ResNet blocks based on the ACDConv and the iterative depth prediction process. Specifically, given an input panoramic color image, the encoder extracts feature maps in five downsampling scales with the ResNet blocks. Here, the $3\times3$ convolution layers are replaced by our proposed ACDConv, the detail is introduced in the following sub-section. Second, the decoder upsamples the feature maps with up-convolution modules \cite{DBLP:conf/3dim/LainaRBTN16} and produces the depth maps in 4 different scales. The first coarse depth map $D_0$ is generated at the 1/8 downsampling level (level $0$), then the subsequent residual maps $\{R_i\}_{i=1}^3$ are produced in the following decoder blocks. The depth map $D_i$ at level $i$ is formulated as $D_i = \tilde{D}_{i-1}+R_i$, where $\tilde{D}_{i-1}$ is up-sampled from $D_{i-1}$ with bilinear interpolation.

\begin{figure}[tb]
\centering
  \includegraphics[width=0.45\textwidth]{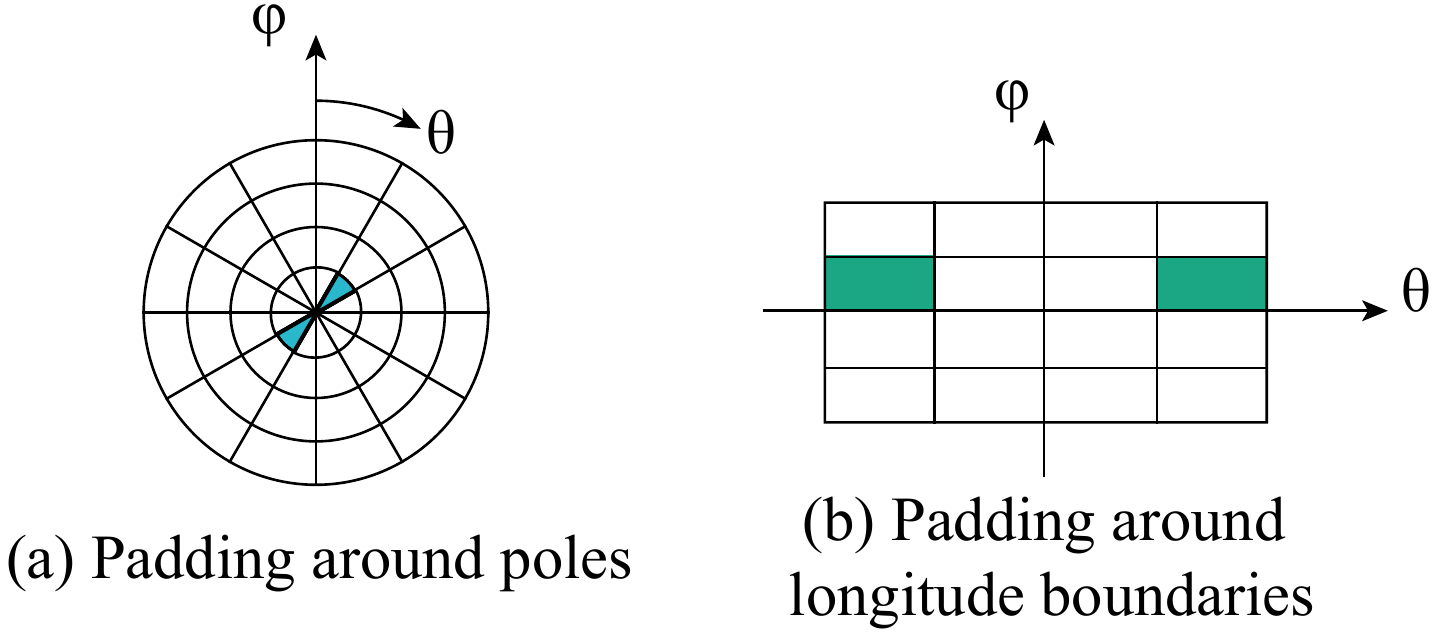}
  \caption{The circular padding for panoramic feature maps. The elements with the same color are selected to pad each other.}
  \label{fig:pad}
\end{figure}

Besides, the circular padding ~\cite{DBLP:conf/icra/WangHLHZS18} is utilized to maintain a complete and continuous spatial field of view for panoramic images. As shown in Fig.\ref{fig:pad}, we select feature items along the longitudinal direction near the poles and the latitudinal direction near the other boundaries. The circular padding maintains the continuity of spatial information on the sphere surface and avoids the invalid padding elements for dilated convolutions.

\subsection{Adaptively Combined Dilated Convolution}

The panorama image expands the spherical imaging result to a rectangular image, which causes the narrow FoV near the poles for regular convolutions. Previous works have developed different custom convolutions to extend FoV near the poles. In this paper, we combine the regular convolutions with different dilations to increase the FoV. Moreover, we introduce an adaptive channel-wise fusion (ACF) module to aggregate the feature maps and get diverse attention areas in the receptive field along the channels. 

\begin{figure}[tb]
\centering
  \includegraphics[width=0.45\textwidth]{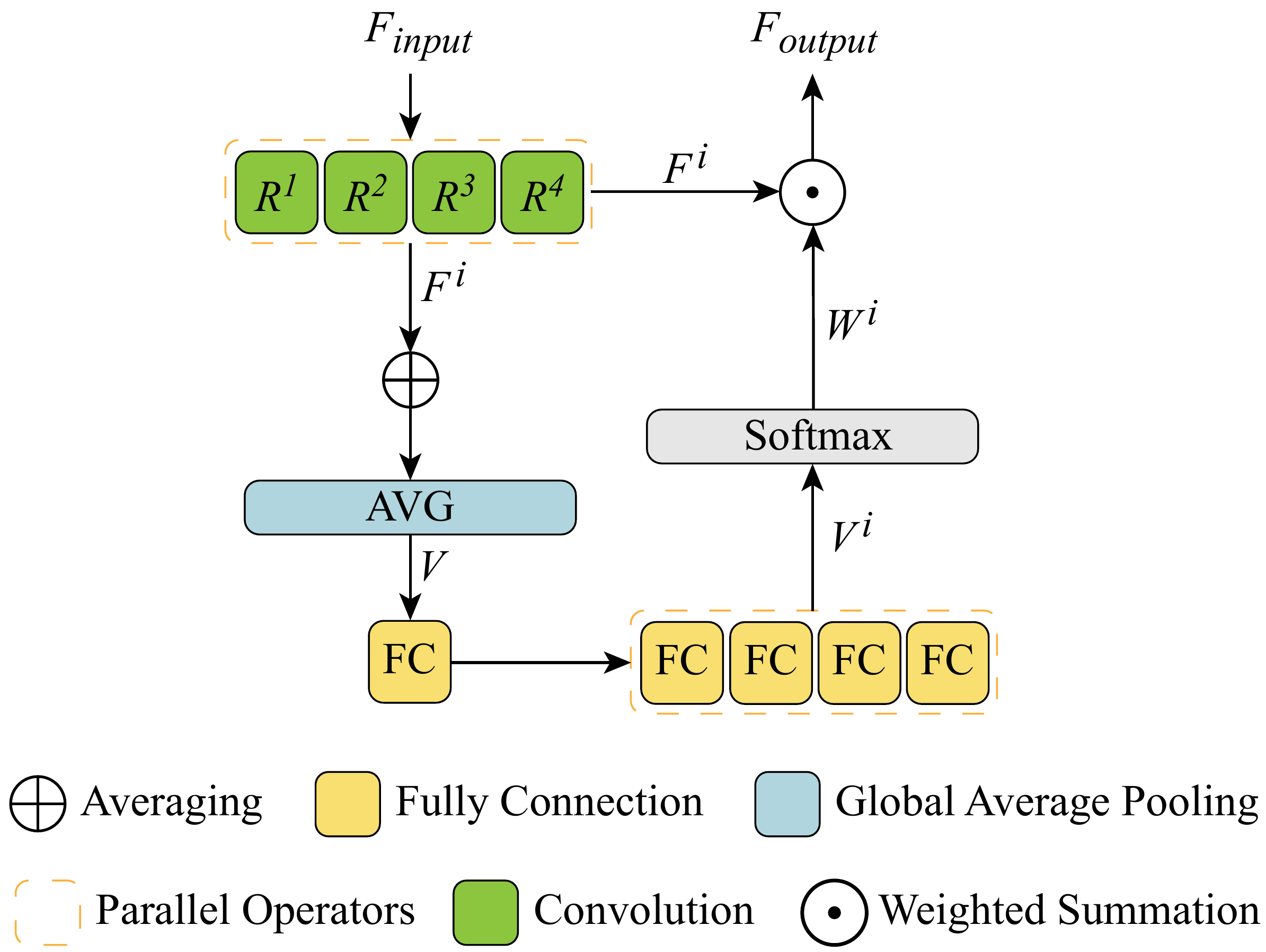}
  \caption{Our adaptively combined dilated convolution layer. The label $R^n$ in the figure means the n-th choice of the four dilation rate settings.}
  \label{fig:convolution}
\end{figure}

The details of our proposed ACDConv are illustrated in Fig.\ref{fig:convolution}. First, given the input features \newchange{$F_{input}\in \mathbb{R}^{C\times H\times W}$}, we use the different convolutions with a group of dilation settings to extract feature maps \newchange{$\{F^i\}_{i=1}^4\in \mathbb{R}^{C'\times H\times W}$} from the input features in parallel. Then, a learnable ACF module is applied to integrate the feature maps. Specifically, we first get the intermediate mean feature and utilize the global average pooling to obtain a vector \newchange{$V\in \mathbb{R}^{C'}$}. \newchange{After that, the fully connected layers predict the probability vectors $\{V^i\}_{i=1}^4\in \mathbb{R}^{C'}$ for different feature maps, and the softmax function is applied to produce the channel-wise fusion weights as follows:
\begin{align}
W^i_c=\frac{\exp{V^i_c}}{\sum^4_{j=1}{\exp{V^j_c}}}
\end{align}
Finally, the feature maps $F^i$ from different convolutions are summarized with the channel-wise weights $W^i$ to generate the final feature maps $F_{output}=\sum_i{F^iW^i}$ and get a large receptive field.}

\begin{figure*}[tb]
\centering
  \includegraphics[width=0.9\textwidth]{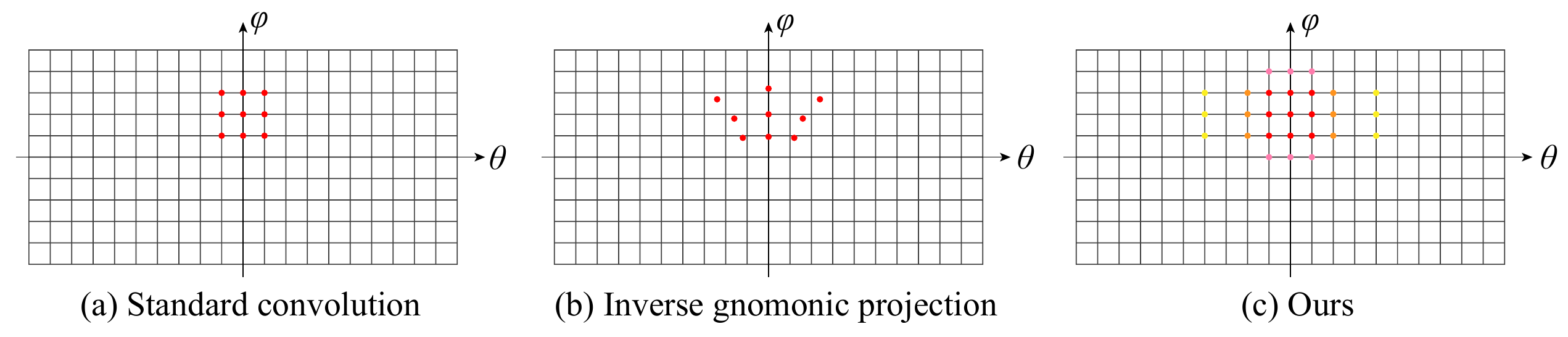}
  \caption{Receptive fields (RF) of different convolutions. In sub-figure (c), the items with the same color in a sub-area of a receptive field have the same interest scores.}
  \label{fig:rf}
\end{figure*}

On the other hand, we draw the receptive fields of different convolutions in Fig.\ref{fig:rf}, including the regular $3\times3$ convolution, the custom convolution with inverse gnomonic projection in SphereNet, and our adaptively combined dilated convolution. As shown in Fig.\ref{fig:rf} (a), the vanilla convolution always keeps the $3\times3$ receptive field in the different latitudes of the ERP, and SphereNet deforms the convolution kernel according to the latitudes to extend the receptive field, especially in the poles areas (see Fig.\ref{fig:rf} (b)). Our combined convolution keeps the shape of a large receptive field in the different latitudes of the ERP as shown in Fig.\ref{fig:rf} (c). More importantly, different attention weights in diverse areas of the combined receptive field along the channels can be acquired after the weighted summarization. Besides, we set the dilation settings as $1\times1$, $1\times2$, $1\times4$, and $2\times1$ in our experiments.

\subsection{Loss Function}
In our approach, we use the BerHu \cite{DBLP:conf/3dim/LainaRBTN16} loss function to supervise the training process for the network, which is formulated as:

\begin{align}
L(d_i,\widetilde{d}_i) = \left\{
    \begin{matrix}
    |\Delta| & |\Delta|\le c \\
    \frac{|\Delta|^2+c^2}{2c} & |\Delta|> c
    \end{matrix}
\right.
\end{align}
where $\Delta=d_i-\widetilde{d}_i$, and $d_i,\widetilde{d}_i$ are the estimated depth and the ground truth on pixel $i$ of input image respectively. 

For each input image, the parameter $c$ is set as
\begin{align}
    c=\frac{1}{5}\max_i |d_i-\widetilde{d}_i|
\end{align}

Finally, we apply the BerHu loss on $D_3$, the shape of which is the same as the input image, and $\{D_i\}_{i=0}^2$ are part of the components of $D_3$.

\section{Experiments}
In this section, we first introduce our experiments, including datasets, implementation details, and evaluation metrics. Then, we provide the qualitative and quantitative comparisons of our network with state-of-the-art approaches.
Finally, we perform the ablation experiments to validate the effectiveness of our network structure.
All experiments were conducted on a server computer equipped with an Intel(R) Xeon(R) Gold 6130 CPU processor, 256GB of RAM, and an NVIDIA TITAN RTX 24GB graphics card.

\begin{table*}[!htb]
\centering
    \begin{tabular}{ccccccccc} 
\toprule 
Dataset & Method & MAE$\downarrow$ & RMSE$\downarrow$ & RMSElog$\downarrow$ &  AbsRel$\downarrow$ & $\delta^1\uparrow$ & $\delta^2\uparrow$ & $\delta^3\uparrow$\\
\midrule 
\multirow{6}{*}{Stanford2D3D} & BiFuse  & 0.2343 & 0.4142 & 0.0787 & 0.1209 & 86.60 & 95.80 & 98.60 \\
                            ~ & UniFuse & 0.2082 & 0.3691 & 0.0721 & 0.1114 & 87.11 & 96.64 & 98.82\\
                            ~ & HoHoNet & 0.2027 & 0.3834 & 0.0668 & 0.1014 & \textbf{90.54} & 96.93 & 98.86\\
                            ~ & SliceNet&  \textbf{0.1757} & 0.3509 & 0.0801 & 0.0995 & 90.29 & 96.26 & 98.44\\
                            ~ & SphereNet & 0.2253 & 0.3833 & 0.0786 & 0.1234 & 85.39 & 95.67 & 98.33\\
                            ~ & Ours    & 0.1870 & \textbf{0.3410} & \textbf{0.0664} & \textbf{0.0984} & 88.72 & \textbf{97.04} & \textbf{98.95} \\

\midrule 
\multirow{6}{*}{Matterport3D} & BiFuse  & 0.3470 & 0.6259 & 0.1134 & 0.2048 & 84.52 & 93.19 & 96.32 \\
                            ~ & UniFuse & 0.2814 & 0.4941 & 0.0701 & 0.1063 & 88.97 & 96.23 & 98.31\\
                            ~ & HoHoNet & 0.2862 &  0.5138 & 0.0871 & 0.1488 & 87.86 & 95.19 & 97.71\\
                            ~ & SliceNet& 0.3296 & 0.6133 &  0.1045 & 0.1764 & 87.16 & 94.83 & 97.16\\
                            ~ & SphereNet & 0.3167 & 0.5212 & 0.0778 & 0.1258 & 84.34 & 95.49 & 98.17 \\
                            ~ & Ours    & \textbf{0.2670} & \textbf{0.4629} & \textbf{0.0646} & \textbf{0.1010} & \textbf{90.00} & \textbf{96.78} & \textbf{98.76}\\
\midrule
\multirow{6}{*}{Structured3D} & BiFuse  & 0.0562 & 0.1100 & 0.0295 & 0.0401 & 98.19 & 99.41 & 99.72 \\
                            ~ & UniFuse & 0.0617 & 0.1167 & 0.0324 & 0.0458 & 97.65 & 99.28 & 99.69 \\
                            ~ & HoHoNet & 0.0549 & 0.1088 & 0.0316 & 0.0408 & 97.97 & 99.35 & 99.70 \\
                            ~ & SliceNet& 0.0660 & 0.1290 & 0.0444 & 0.0496 & 97.25 & 99.09 & 99.54\\
                            ~ & SphereNet & 0.0664 & 0.1161 & 0.0368 & 0.0491 & 97.58 & 99.36 & 99.71 \\
                            ~ & Ours    & \textbf{0.0454} & \textbf{0.0924} & \textbf{0.0291} & \textbf{0.0327} & \textbf{98.74} & \textbf{99.59} & \textbf{99.82}\\
\bottomrule
    \end{tabular}
    \caption{Quantitative comparison on different datasets. The best result of each measurement is marked in \textbf{bold} font. Here we re-train the previous works on Structured3D with their source codes at the resolution of $512\times1024$, and the training strategies are the same as ours.}
    \label{tab:comparison-1}
\end{table*}

\subsection{Implementation}
\subsubsection{Datasets}

We carry out experiments on both virtual and real-world datasets, including Stanford2D3D \cite{DBLP:journals/corr/ArmeniSZS17}, Matterport3D \cite{DBLP:conf/3dim/ChangDFHNSSZZ17}, and Structured3D \cite{DBLP:conf/eccv/ZhengZLTGZ20}. Both Stanford2D3D and Matterport3D are scanned with RGB-D cameras in the real-world scenes, and they include $1,413$ and $10,800$ RGB-D panoramic views respectively. While Structured3D is rendered with synthetic scenes, and it contains over $196k$ RGB-D panorama images. 
For Stanford2D3D and Matterport3D, we follow their official splits with entire panoramic RGB-D pairs to train and test the network. For Structured3D, we just utilize the subset with \textit{rawlight} illumination and \textit{full} furniture settings. The subset includes $21,835$ panoramic RGB-D image pairs, and we follow the official scene split for training and testing. 
Moreover, we follow the process strategy for Matterport3D as previous works to merge the 18-views perspective depth images and the rendered skybox color images to panoramic RGB-D image pairs.

\subsubsection{Implementation Details}
We implement our network on the PyTorch \cite{DBLP:conf/nips/PaszkeGMLBCKLGA19} platform. We train our network for 100 epochs on Stanford2D3D, 60 epochs on Matterport3D, and 60 epochs on Structured3D with Adam \cite{DBLP:journals/corr/KingmaB14} optimizer respectively, the learning rate is set as 1e-4 in all the experiments. 
Meanwhile, we set the image size as $512\times1024$ with the batch size of 6 on an NVIDIA TITAN RTX graphics card.

\subsubsection{Evaluation Metrics}
We adopt five widely-used evaluation metrics used in previous works to evaluate our method quantitatively, including mean absolute error (MAE), root mean square error (RMSE), logarithmic root mean square error (RMSElog), absolute relative error (Abs Rel), and threshold percentage ($\delta^n$), which can be formulated as:
\begin{itemize}
    \item[\textbf{$\cdot$}] $ MAE=\frac{1}{N}\sum_{i=1}^N |d_i-\widetilde{d}_i| $;
    \item[\textbf{$\cdot$}] $ RMSE=\sqrt{\frac{1}{N}\sum_{i=1}^N |d_i-\widetilde{d}_i|^2} $;
    \item[\textbf{$\cdot$}] $ RMSElog=\sqrt{\frac{1}{N}\sum_{i=1}^N |\log d_i - \log \widetilde{d}_i|^2}$;
    \item[\textbf{$\cdot$}] $ Abs Rel=\frac{1}{N}\sum_{i=1}^N |d_i-\widetilde{d}_i|/\widetilde{d}_i$;
    \item[\textbf{$\cdot$}] Threshold percentage $\delta^n$ is the percentage of pixels satisfying $\max(\frac{d_i}{\widetilde{d}_i},\frac{\widetilde{d}_i}{d_i})<1.25^n$.
\end{itemize}
Following the previous methods, we clip the estimated depth maps to $10m$ without scale calibration when calculating the evaluation metrics.

\subsection{Comparison Experiments}

In this sub-section, we provide the quantitative comparison and visual comparison to prove the effectiveness of our method.
% Besides, we also present the comparison of reconstructed point clouds in our supplementary material.

\subsubsection{Quantitative Comparison}

\newchange{
We compare our ACDNet with previous works on the three above-mentioned datasets, and the quantitative results are shown in Tab.\ref{tab:comparison-1}. Our ACDNet outperforms previous works for most metrics on Stanford2D3D and all metrics on Matterport3D and Structured3D. Note that the results of SliceNet on Stanford2D3D are produced by the fixed parameters in SliceNet's Github repository \footnote{https://github.com/crs4/SliceNet} and differ from the original values in its paper. Specifically, our results exceed the previous state-of-the-art method by $5.1\%$ in MAE metric on Matterport3D and $17.3\%$ on Structured3D as well as reduce the AbsRel metric by $1.1\%$, $5.0\%$, and $18.5\%$ on the three datasets. Besides, we also compare our method with the SphereNet that uses a custom convolution, where we implement by using the same framework and replace the convolution layers in ResNet50 with that in SphereNet. According to the results in Tab.\ref{tab:comparison-1}, our ACDNet with ACDConv also outperforms the distortion-aware convolution with inverse gnomonic projection in SphereNet. On the one hand, the ACDConv expands the receptive field to get a large spatial FoV. On the other hand, it focuses on the various areas in the receptive field along different channels, which makes the convolution kernels learn a variety of latent kernel shapes in different channels, and our network can accommodate the diverse geometric relationships in ERP.
}

% We compare our ACDNet with previous works on the three above-mentioned datasets, and the quantitative results are shown in Tab.\ref{tab:comparison-1}. Previous SOTA work on Stanford2D3D, i.e., SliceNet, performs poorly on the other two datasets, while our approach achieves the second-best performance on Stanford2D3D and the best performance on Matterport3D and Structured3D. Specifically, our results exceed the previous state-of-the-art method by $5.1\%$ in MAE metric on Matterport3D and $17.3\%$ on Structured3D. Besides, we also compare our method with the SphereNet that uses a custom convolution, where we implement by using the same framework and replace the convolution layers in ResNet50 with that in SphereNet. According to the results in Tab.\ref{tab:comparison-1}, our ACDNet with ACDConv also outperforms the distortion-aware convolution in SphereNet. On the one hand, the ACDConv expands the receptive field to get a large spatial FoV. On the other hand, it focuses on the various areas in the receptive field along different channels, which makes the convolution kernels learn a variety of latent kernel shapes in different channels, and our network can accommodate the diverse geometric relationships in ERP.

\begin{figure*}[tb]
\centering
  \includegraphics[width=1.0\textwidth]{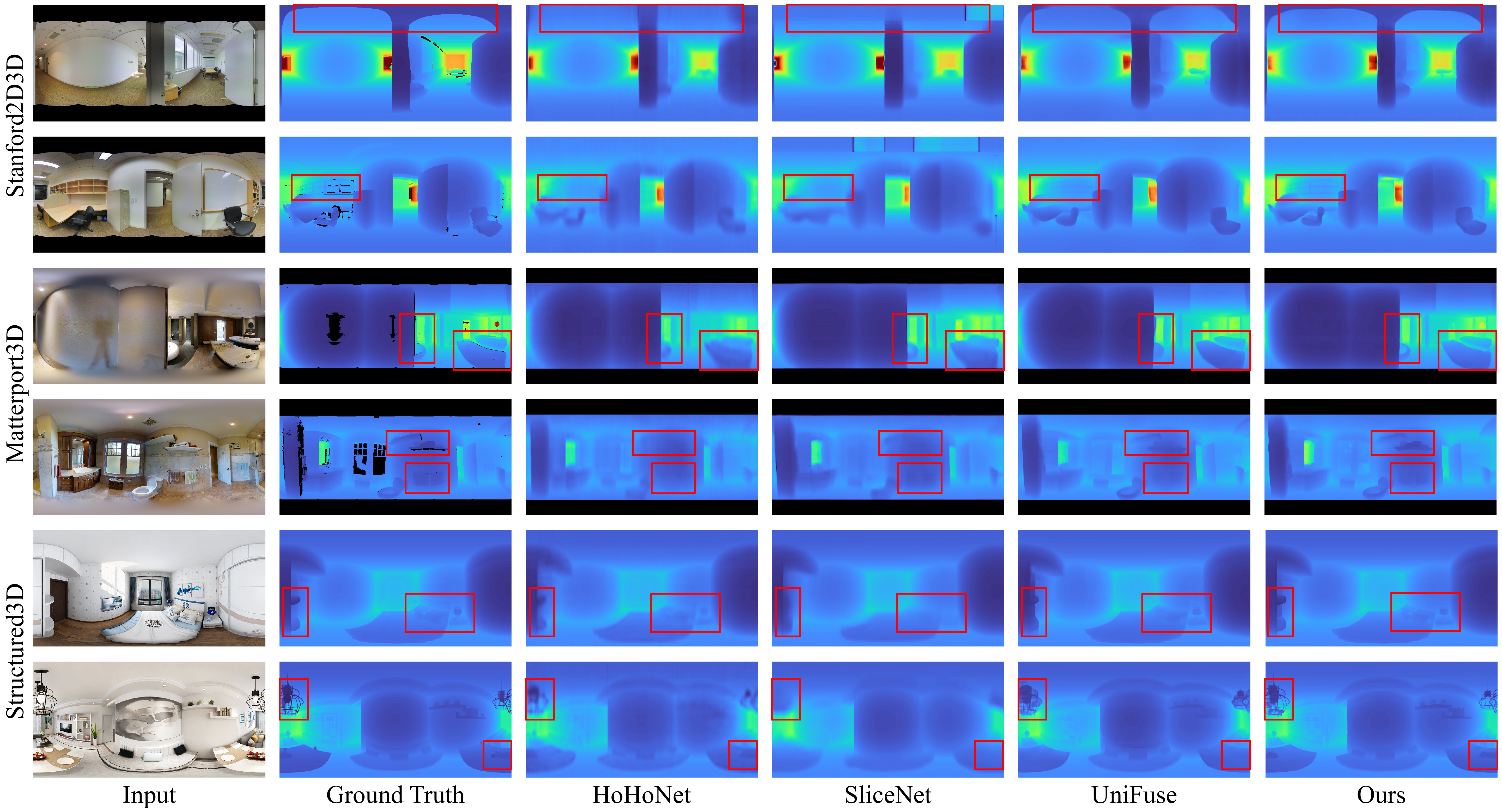}
  \caption{Depth maps comparison with other methods. The area with zero values in ground truth means the missing area of depth maps.}
  \label{fig:visual}
\end{figure*}

\subsubsection{Visual Comparison}
Furthermore, Fig.\ref{fig:visual} shows our visual comparison results with SOTA methods on three datasets.
As shown in Fig.\ref{fig:visual}, our ACDNet predicts more accurate and detailed depth maps with better visual effects. Firstly, we recover more clear and accurate walls in the invisible areas, as shown in the first row of Fig.\ref{fig:visual}. It can be easily noticed that our ACF module plays a key role in capturing the spatial global context and the circular padding helps to keep the spatial continuity of the panorama images on the sphere.
Compared to our single-branch, the cubemap branch proposed in the UniFuse extracts features with weak texture in the ceiling areas of a narrow FoV, which leads to bad performance in the obscured areas. Secondly, our ACDNet estimates more object details in the depth maps, such as the bookcase in the second row and the bathtub in the third row in Fig.\ref{fig:visual}. Our network also performs well in distinguishing between the background and the foreground objects with similar depth values as shown in the last three rows in Fig.\ref{fig:visual}. Moreover, our ACDNet generates more accurate edges in the depth maps, such as the wall in the third row and the ceiling lamps in the last row in Fig.\ref{fig:visual}. These results demonstrate the better performance of our ACDNet in the depth maps estimation with monocular panorama images.

\subsection{Ablation Studies}

To further verify the effectiveness of our ACDNet, we introduce some groups of ablation studies on the Stanford2D3D dataset in this section. First, we conduct some ablation studies on our ACDConv, including the different parts, the different dilation directions, and the number of dilated convolutions. 
\newchange{
Then, we compare the results of different padding methods and study the advantage of iterative depth prediction.
Finally, we test the network with different ResNet backbones. In addition, we also compare the model complexity and inference time with existing methods. In all of these experiments, we use the same hyper-parameters and training strategy on Stanford2D3D.
}

\subsubsection{Adaptively Combined Dilated Convolution}
Here, four experiments are executed to study the roles of the dilated convolution and the ACF module in our ACDConv as shown in Tab.\ref{tab:abs-1}. 
First, we remove the ACDConv and use the original ResNet backbone with the regular convolution in our network as the baseline. Then we introduce the ACDConv but replace our ACF module with a simple average operator, denoted as \textit{Simple}.
% Moreover, we apply the adaptive row-wise fusion strategy in ACDConv instead of the channel-wise fusion strategy, denoted as \textit{Row-wise}. Specifically, given the intermediate features with the shape of $B\times C\times H\times W$, we squeeze the features in dimensions of $C$ and $W$ with an averaging operator to get a vector of $B\times H$, then the following process is similar to the channel-wise vectors shown in Fig.\ref{fig:convolution}. 
\newchange{
Moreover, we test two other fusion strategies in ACDConv instead of our channel-wise fusion strategy. Specifically, given the intermediate feature maps $\{F^i\}_{i=1}^4\in \mathbb{R}^{C'\times H\times W}$, the \textit{Row-wise} strategy adds $F^i$ and squeezes the results to row-wise feature vector $V\in \mathbb{R}^{H}$ with averaging operators, then the following MLPs generate the row-wise normalized probability $\{W^i\}_{i=1}^4 \in \mathbb{R}^{H}$ for fusion process. By contrast, the \textit{Pixel-wise} strategy does not squeeze the feature maps and produces pixel-wise normalized probability $\{W^i\}_{i=1}^4 \in \mathbb{R}^{C'\times H\times W}$ with $1\times1$ convolutions.
}
Finally, we also test the popular ASPP module in our baseline which combines the different dilated convolutions to capture multi-scale context. 

\begin{table}[tb]
    \centering
    % \small
    % \setlength\tabcolsep{3pt}
    \begin{tabular}{ccccc} 
\toprule 
Method & MAE$\downarrow$ & RMSE$\downarrow$ & RMSElog$\downarrow$ &  AbsRel$\downarrow$ \\
\midrule 
Baseline  & 0.2104 & 0.3620 & 0.0746 & 0.1148 \\
Simple & 0.2037 & 0.3582 & 0.0689 & 0.1075 \\
Row-wise & 0.2096 & 0.3694 & 0.0759 & 0.1124 \\
Pixel-wise & 0.2096 & 0.3659 & 0.0720 & 0.1090 \\
Ours   & \textbf{0.1870} & \textbf{0.3410} & \textbf{0.0664} & \textbf{0.0984} \\
ASPP & 0.1990 & 0.3633 & 0.0700 & 0.1032 \\
\bottomrule
    \end{tabular}
    \caption{Ablation studies about ACDConv. In the table, we use the different convolutions in the ResNet50 blocks to build the network.}
    \label{tab:abs-1}
\end{table}

As shown in Tab.\ref{tab:abs-1}, simply averaging the features from different dilated convolutions or utilizing the ASPP module can improve the performance on panoramic images, which shows that the large receptive field of convolution contributes to this task. 
Our adaptive aggregation operator produces various attention areas on the receptive field and can adapt to the different geometric properties in the different latitudes of panoramic images, thus our full ACDConv outperforms these two methods.
\newchange{
Besides, the adaptively combined dilated convolution with row-wise or pixel-wise fusion strategy hardly improves the results or even makes it worse. 
Thus, explicitly combining different dilation at different latitudes or different pixel locations is not suitable for panoramic images and our adaptively learning the receptive field in different feature channels is a more applicable solution, which also explains the reason that the custom distortion-aware convolution does not perform well in this task.
}

\begin{table}[tb]
    \centering
    % \small
    % \setlength\tabcolsep{3pt}
    \begin{tabular}{ccccc} 
\toprule 
Method & MAE$\downarrow$ & RMSE$\downarrow$ & RMSElog$\downarrow$ &  AbsRel$\downarrow$ \\
\midrule 
Baseline  & 0.2104 & 0.3620 & 0.0746 & 0.1148 \\
X-axis & 0.1953 & 0.3559 & 0.0679 & 0.1028 \\
Y-axis & 0.1965 & 0.3540 & 0.0684 & 0.1028 \\
Ours   & \textbf{0.1870} & \textbf{0.3410} & \textbf{0.0664} & \textbf{0.0984} \\
\bottomrule
    \end{tabular}
    \caption{Ablation studies about different dilation directions.}
    \label{tab:abs-2}
\end{table}

Due to the deployment of the dilation convolutions along both the x-axis and the y-axis, we study the effects of different dilation directions. 
Specifically, we separately test the dilation only along the x-axis or the y-axis with the combined dilation settings of 1, 2, 3, and 4. 
As shown in Tab.\ref{tab:abs-2}, dilation settings along only the x-axis or y-axis can improve the MAE metric to 0.1953 and 0.1965 respectively. 
As the areas near the poles have narrow FoV along the latitude direction in the ERP coordinates, the dilated convolution along the latitude contributes to addressing this problem. 
Moreover, despite that the ERP expression keeps uniform FoV along the longitude direction, the areas near the poles mainly include the weak textures and geometric structures, e.g., the ceiling and the floor. Thus, the dilation along the y-axis introduces more spatial information in these areas in estimating accurate depth. 
Therefore, our dilation settings along both directions have the best performance (see in Tab.\ref{tab:abs-2}).

We also discuss the impact of the dilation number in the ACDConv shown in Tab.\ref{tab:abs-4}. More dilations could bring larger receptive fields and better depth estimation performance but increase the network complexity and aggravate training difficulty at the same time. 
When the dilation number is 5 with an additional $1\times8$ dilation setting, the network is overfitted, which will make performance worse.
Thus, the dilation number is set to 4 in our experiments.

\begin{table}[tb]
    \centering
    % \small
    % \setlength\tabcolsep{3pt}
    \begin{tabular}{ccccc} 
\toprule 
Dilations & MAE$\downarrow$ & RMSE$\downarrow$ & RMSElog$\downarrow$ &  AbsRel$\downarrow$ \\
\midrule 
Baseline  & 0.2104 & 0.3620 & 0.0746 & 0.1148 \\
Two & 0.2038 & 0.3632 & 0.0703 & 0.1088 \\ % MAE:0.2038 | RMSE:0.3632 | LogRMSE:0.0703 | Abs:0.1088 | D1:0.8747 | D2:0.9664 | D3:0.9884 |
Three & 0.1971 & 0.3573 & 0.0687 & 0.1023 \\ % MAE:0.1971 | RMSE:0.3573 | LogRMSE:0.0687 | Abs:0.1023 | D1:0.8752 | D2:0.9678 | D3:0.9901 |
Four (Ours)  & \textbf{0.1870} & \textbf{0.3410} & \textbf{0.0664} & \textbf{0.0984} \\
Five & 0.1963 & 0.3561 & 0.0689 & 0.1047 \\ 
\bottomrule
    \end{tabular}
    \caption{Ablation studies about different dilation numbers.}
    \label{tab:abs-4}
\end{table}

\subsubsection{Padding Method}

In the proposed ACDNet, we apply circular padding to get continuous features on the sphere. In this sub-section, we also test the effects of zero padding and left-right padding as shown in Tab.\ref{tab:abs-3}. We observe that the results have gradual improvements by zero padding, left-right padding, and circular padding. 
\newchange{
The root cause is that proper padding avoids introducing abundant invalid elements into the dilated convolutions in the boundary regions.
}
Meanwhile, this also demonstrates the effectiveness of the complete and continuous spatial information for depth estimation in the panoramic images.

\begin{table}[tb]
    \centering
    \begin{tabular}{ccccc} 
\toprule 
Padding & MAE$\downarrow$ & RMSE$\downarrow$ & RMSElog$\downarrow$ &  AbsRel$\downarrow$ \\
\midrule 
ZeroPad & 0.1948 & 0.3526 & 0.0684 & 0.1045 \\
LRPad & 0.1935 & 0.3503 & 0.0670 & 0.1025 \\
CirPad & \textbf{0.1870} & \textbf{0.3410} & \textbf{0.0664} & \textbf{0.0984} \\
\bottomrule
    \end{tabular}
    \caption{Ablation studies about different padding methods.}
    \label{tab:abs-3}
\end{table}

\begin{table}[tb]
    \centering
    \begin{tabular}{ccccc} 
\toprule 
Methods & MAE$\downarrow$ & RMSE$\downarrow$ & RMSElog$\downarrow$ &  AbsRel$\downarrow$ \\
\midrule 
\small{Baseline} & \multirow{2}{*}{0.2104} & \multirow{2}{*}{0.3620} & \multirow{2}{*}{0.0746} & \multirow{2}{*}{0.1148} \\
\small{w/ iter} & ~ & ~ & ~ & ~\\
\small{Baseline} & \multirow{2}{*}{0.2287} & \multirow{2}{*}{0.3923} & \multirow{2}{*}{0.0796} & \multirow{2}{*}{0.1200}\\
\small{w/o iter} & ~ & ~ & ~ & ~\\
\small{Ours}   & \multirow{2}{*}{\textbf{0.1870}} & \multirow{2}{*}{\textbf{0.3410}} & \multirow{2}{*}{\textbf{0.0664}} & \multirow{2}{*}{\textbf{0.0984}} \\
\small{w/ iter} & ~ & ~ & ~ & ~\\
\small{Ours} & \multirow{2}{*}{0.2017} & \multirow{2}{*}{0.3650} & \multirow{2}{*}{0.0694} & \multirow{2}{*}{0.1022} \\
\small{w/o iter} & ~ & ~ & ~ & ~\\
\bottomrule
    \end{tabular}
    \caption{Ablation studies about iterative depth prediction.}
    \label{tab:abs-5}
\end{table}

\newchange{
\subsubsection{Iterative Depth Prediction}
In this sub-section, we separately test the role of iterative depth prediction and our ACDConv.
%the role of iterative depth prediction is tested separately from our ACDConv. 
According to Tab.\ref{tab:abs-5}, iterative depth prediction improves the performance in both baseline and our network as it decomposes different scales of depth regression and improves the process of gradient backpropagation. 
Meanwhile, our ACDConv efficiently extracts features for more precise depth estimation and works independently with iterative depth prediction.
}

% \begin{table}[!htb]
%     \centering
%     \begin{tabular}{ccccc} 
% \toprule 
% Methods & MAE$\downarrow$ & RMSE$\downarrow$ & RMSElog$\downarrow$ &  AbsRel$\downarrow$ \\
% \midrule 
% \small{Baseline} & \multirow{2}{*}{0.2287} & \multirow{2}{*}{0.3923} & \multirow{2}{*}{0.0796} & \multirow{2}{*}{0.1200}\\
% \small{w/o iter} & ~ & ~ & ~ & ~\\
% Baseline & 0.2104 & 0.3620 & 0.0746 & 0.1148 \\
% \small{Full} & \multirow{2}{*}{0.2017} & \multirow{2}{*}{0.3650} & \multirow{2}{*}{0.0694} & \multirow{2}{*}{0.1022} \\
% \small{w/o iter} & ~ & ~ & ~ & ~\\
% Full   & \textbf{0.1870} & \textbf{0.3410} & \textbf{0.0664} & \textbf{0.0984} \\
% \midrule
% \bottomrule
%     \end{tabular}
%     \caption{Ablation studies about iterative depth prediction.}
%     \label{tab:abs-5}
% \end{table}

% \begin{table}[!htb]
%     \centering
%     \begin{tabular}{ccccc} 
% \toprule 
% Methods & MAE$\downarrow$ & RMSE$\downarrow$ & RMSElog$\downarrow$ &  AbsRel$\downarrow$ \\
% \midrule 
% Baseline woi & 0.2287 & 0.3923 & 0.0796 & 0.1200\\
% Baseline & 0.2104 & 0.3620 & 0.0746 & 0.1148 \\
% Full woi & 0.2017 & 0.3650 & 0.0694 & 0.1022 \\
% Full   & \textbf{0.1870} & \textbf{0.3410} & \textbf{0.0664} & \textbf{0.0984} \\
% \midrule
% \bottomrule
%     \end{tabular}
%     \caption{Ablation studies about iterative depth prediction.}
%     \label{tab:abs-5}
% \end{table}

\subsubsection{ResNet Backbone}
Finally, we test different ResNet backbones in our ACDNet. As shown in Tab.\ref{tab:abs-6}, the network performance gradually improves with the increasing of the backbone complexity. However, using the ResNet101 backbone to build the network is time-consuming and produces overfitting. Considering network performance and overhead, we select ResNet50 as our backbone in the experiments.

\begin{table}[tb]
    \centering
    \begin{tabular}{ccccc} 
\toprule 
Backbone & MAE$\downarrow$ & RMSE$\downarrow$ & RMSElog$\downarrow$ &  AbsRel$\downarrow$ \\
\midrule 
ResNet18 & 0.2309 & 0.3957 & 0.0771 & 0.1195 \\
ResNet34 & 0.2044 & 0.3661 & 0.0687 & 0.1041 \\
ResNet50 & \textbf{0.1870} & \textbf{0.3410} & 0.0664 & \textbf{0.0984} \\
ResNet101 & 0.1911 & 0.3481 & \textbf{0.0654} & 0.0992 \\ 
\bottomrule
    \end{tabular}
    \caption{Ablation studies about different ResNet backbones.}
    \label{tab:abs-6}
\end{table}

\newchange{
\subsubsection{Model Complexity}
We compare the model complexity and computational efficiency with previous methods, and all the results are derived from inferring a $512\times 1024$ image.
Compared with \textit{Baseline}, the dilated convolution groups introduced in \textit{Simple} increase $33.9M$ parameters and $240M$ memories and reduce the FPS from 19 to 12 as shown in Tab.\ref{tab:abs-7}.
Against the data in Tab.\ref{tab:abs-1}, simply increasing model parameters does not play a fundamental role in improving the performance. While the channel-wise fusion modules in \textit{Ours} scarcely influence the model complexity and computational efficiency but improve results substantially.

\begin{table}[tb]
\centering
\small
\begin{tabular}{ccccc} 
\toprule 
Method & Parameters & GPU Mem & FPS\\
\midrule 
BiFuse & 253.1M & 4003M & 0.9 \\
UniFuse & 30.26M & 1221M & 31 \\
SliceNet & 75.3M & 1911M & 13 \\
HoHoNet & 49.5M & 1487M & 52 \\
Ours (Baseline)  & 52.5M & 2136M & 19 \\
Ours (Simple) & 86.4M & 2376M & 12 \\
Ours & 87.0M & 2378M & 11 \\
\bottomrule
\end{tabular}
\caption{Model complexity and computation efficiency}
\label{tab:abs-7}
\end{table}

}

\section{Conclusion}
In this work, we first propose the adaptively combined dilated convolution to replace the regular convolution to well extract the features from panorama images in ERP. Then we construct the ACDNet to estimate depth maps with monocular panorama images, which outperforms the SOTA approaches in quantitative metrics and visual effects. 

Our experiments show that the convolutions with extended receptive fields contribute to panoramic depth estimation. Moreover, the experiments with adaptive channel-wise fusion strategy also express that obtaining different latent shapes of convolution kernels in different feature channels is better than explicitly deforming convolution kernels at different latitudes. That is worth further researching for panorama images in the future.
\newchange{
In addition, we will study our ACDConv in other existing depth prediction models and its effects on other various panoramic image tasks such as image classification and semantic segmentation.
}
\section{Acknowledgments}

% This work is supported by the Strategic Priority Research Program of the Chinese Academy of Sciences (No.XDA23090304), the National Natural Science Foundation of China (U2003109), the Key Research Program of Frontier Sciences CAS (QYZDY-SSW-SYS004), the Youth Innovation Promotion Association of the Chinese Academy of Sciences (Y201935), and the Fundamental Research Funds for the Central Universities.
This work is supported by the Strategic Priority Research Program of the Chinese Academy of Sciences (No. XDA23090304), the National Natural Science Foundation of China (U2003109, U21A20515, 62102393), the Youth Innovation Promotion Association of the Chinese Academy of Sciences (Y201935), the State Key Laboratory of Robotics and Systems (HIT) (SKLRS-2022-KF-11), and the Fundamental Research Funds for the Central Universities.
\bibliography{main} 

\end{document}